# Real-Time Lane Detection via Efficient Feature Alignment and Covariance Optimization for Low-Power Embedded Systems

Yian Liu[1] · Xiong Wang[2] · Ping Xu[1], Lei Zhu[1] · Ming Yan[1] · Linyun Xue[1]

**Abstract**
Real-time lane detection in embedded systems encounters significant challenges due to subtle and sparse visual signals in RGB images, often constrained by limited computational resources and power consumption. Although deep learning models for lane detection―categorized into segmentation-based, anchor-based, and curve-based methods―there remains a scarcity of universally applicable optimization techniques tailored for low-power embedded environments. To overcome this, we propose an innovative Covariance Distribution Optimization (CDO) module specifically designed for efficient, real-time applications. The CDO module aligns lane feature distributions closely with ground-truth labels, significantly enhancing detection accuracy without increasing computational complexity. Evaluations were conducted on six diverse models across all three method categories, including two optimized for real-time applications and four state-of-the-art (SOTA) models, tested comprehensively on three major datasets: CULane, TuSimple, and LLAMAS. Experimental results demonstrate accuracy improvements ranging from 0.01% to 1.5%. The proposed CDO module is characterized by ease of integration into existing systems without structural modifications and utilizes existing model parameters to facilitate ongoing training, thus offering substantial benefits in performance, power efficiency, and operational flexibility in embedded systems.

## 1 Introduction

Lane detection or recognition is a critical task in the field of self-driving and advanced driver assistance systems (ADAS), as well as in broader autonomous robotics applications. Accurate lane perception facilitates decision-making processes, provides lane departure warning (LDW), supports adaptive cruise control (ACC), and supplies navigational guidance for collision avoidance [1]. Beyond automotive systems, this technology extends to autonomous robots operating in structured or semi-structured environments, enabling robust path following, route planning, and safe navigation [2].

The lane detection problem is extremely difficult in practice due to several issues, such as thin lane structures, damaged or disappearing edge boundaries, obstruction by other vehicles, incorrect road markers, and poor lighting conditions [3].

✉Yian Liu
   yaliu@hdu.edu.cn
   Xiong Wang
   fanyway@foxmail.com
   Ping Xu
   xuping@hdu.edu.cn
   Lei Zhu
   zhulei@hdu.edu.cn
   Ming Yan
   yanming@hdu.edu.cn
   Linyun Xue
   xly@hdu.edu.cn

[1] School of Automation, Hangzhou Dianzi University, Hangzhou, China

[2] Hangzhou Taikafo tech. Co. Ltd, Hangzhou, China

Additionally, for deployment on mobile devices, the real-time processing effectiveness of images and the lightweight needs of models are crucial [4]. Unlike CNN-based lane detection methods, the latest research proposes many new network structures to extract high-level lane information, such as line shape, curve parameters, and high-level semantic features of lanes from global context information [5]. Although the state-of-the-art (SOTA) network framework has achieved significant advantages over traditional methods, there is still a lack of general methods to optimize network performance across all three categories. In practical applications, a general optimization method should not increase the inference time and should be able to suppress lane false predictions by extracting global context information [6].

In this paper, we propose a Covariance Distribution Optimization (CDO) module tailored for lightweight configurations commonly adopted in embedded systems, where low-power consumption is crucial. By calculating the covariance matrices between lane features and ground-truth labels in both vertical and horizontal directions, the CDO module achieves efficient feature alignment for real-time lane detection. This design is inspired by Variance-Invariance-Covariance Regularization (VICReg), which achieves state-of-the-art performance in self-supervised methods for image representation learning [7]. The covariance matrix is computed through matrix multiplication and is utilized as a regional similarity descriptor between lane features and ground-truth labels. The diagonal elements of the matrix represent self-similarity along corresponding directions, while the off-diagonal elements capture cross-directional correlations, offering a comprehensive characterization of spatial feature alignment. The covariance matrix can represent the similarity between the lane features and the actual lane locations, so increasing the similarity can effectively improve the lane detection results. As shown in Fig. 1, the CDO module compares the similarity between feature maps and ground truth (GT) label masks along two orthogonal directions, horizontal and vertical. Since the correlation algorithm considers the long-distance relationship between pixels, this method can overcome the problem of signal sparseness in lane detection while maintaining the overall geometry feature of the lane. In this way, the position-related information of every lane could be reinforced in the feature map, and that could improve the end-to-end lane detection accuracy. The three key benefits of CDO are as follows: first, CDO is simple to implement after any feature map, provided that the lanes are not too close to one another; second, most lane detection models' total loss functions can readily be supplemented with the CDO-based loss function, and CDO does not require redesigning the original structures of the network; third, CDO can use the existing parameters of the model to optimize the original model, and it won't increase the time of lane inference.

The main contributions of this paper can be summarized as follows:
- We propose a lightweight and generalizable Covariance Distribution Optimization (CDO) module that can be seamlessly integrated into a wide range of state-of-the-art (SOTA) lane detection models, including those optimized for real-time and embedded applications;
- By leveraging covariance computation and diagonal convex estimation, we demonstrate that optimizing the distribution of covariance matrices between lane features and label masks significantly improves the performance of existing lane detection frameworks;
- The proposed CDO module enables model training using existing parameters and can be deployed in current lane detection systems without modifying the model architecture, thereby offering an efficient and low-power optimization solution suitable for embedded systems.

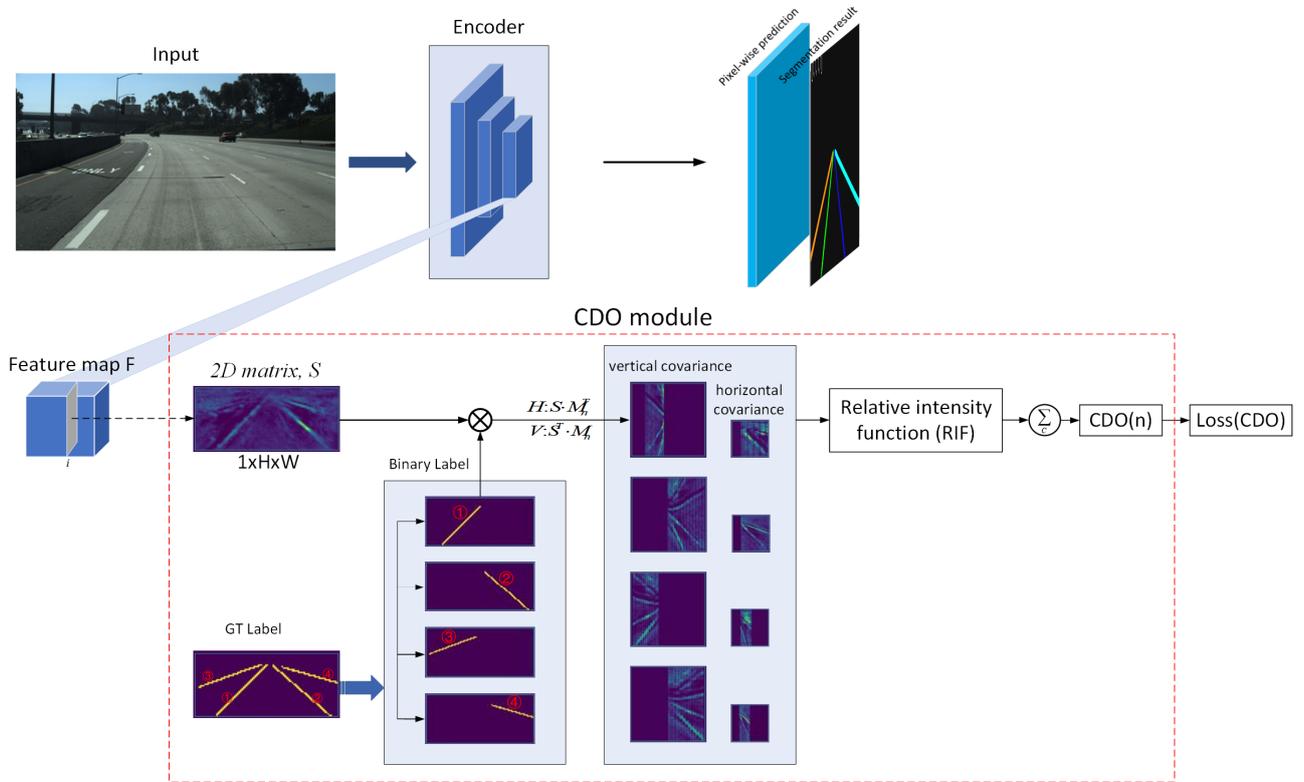

**Fig. 1** The architecture design of CDO, the CDO module, is based on covariance calculation and RIF. The workflow of CDO is as follows: the feature map and the label mask of each lane are multiplied to produce a vertical covariance matrix and a horizontal covariance matrix for that lane. These matrices are then used in RIF calculation. CDO(n) uses RIF to calculate the similarity of lane features and labels in two orthogonal directions. The loss of CDO is used to optimize the original lane detection network.

## 2 Related work

Traditional lane detection techniques have various drawbacks, such as the complex hand-crafted feature selection procedure that is only appropriate for straightforward driving scenes with normal lighting and minor occlusions [8]. Recently, deep learning based methods have shown superiority in lane detection in an end-to-end way [9]. These approaches usually treat lane detection in three categories: segmentation-based method, anchor-based method, and curve-based method [10].

### 2.1 Segmentation-based lane detection

Segmentation-based methods usually treat lane detection as a per-pixel segmentation task. These methods typically combine two separate modules: the encoder and the decoder [11]. The encoder module uses a series of convolutional layers to extract deep features. The decoder module recovers the spatial details from the features. Many recently published segmentation-based methods attempt to modify the encoder to increase lane detection precision. SCNN employs a message-passing mechanism in the hidden layer of the encoder to aggregate spatial information in discrete lane pixels [12]. LaneNet proposes a binary segmentation branch and an instance segmentation branch in the encoder to ensure the lane detection method is robust against road plane changes [13]. RESA proposes a real-time recurrent feature-shift aggregator, which enables the network to gather the global feature and improve performance [14]. The adaptive multi-lane detection method proposes a method that combines instance segmentation with lane center estimation, which can extract discriminative features for foreground extraction [15].

### 2.2 Anchor-based lane detection

The anchor-based model treats lane detection as a point-based object detection problem and uses anchors for feature pooling. Line-CNN uses predefined first-level line anchors in lane detection. The anchor structure converts region-based object representations in Faster R-CNN to line-based representations [16]. LaneATT proposes a single-stage attention mechanism based on line anchors, and the feature pooling operation is similar to Fast R-CNN [17]. It achieves state-of-the-art results and shows both higher efficacy and efficiency. UFAST formulates lane detection as a row anchor-based classification problem. For each row, UFAST predicts the locations of lanes and adopts lightweight backbones to achieve high inference speed [18]. CondLaneNet proposes a recurrent instance module (RIM) to overcome the problem of complex lane topologies such as dense lines and fork lines [19]. CondLaneNet performs row anchor-based formulation and predicts the line shape conditioned on the dynamic kernel parameters. CLRNet adopts the FPN structure to obtain the lane prior features and proposes the ROIGather module to capture the global context information [20].

### 2.3 Curve-based lane detection

Curve-based methods treat lanes as polynomial curves and regress curve parameters to lanes. PolyLaneNet adopts a polynomial representation for the lanes and predicts the polynomial coefficients with fully-connected layers [21]. LSTR introduces the Transformer into the lane detection task, which predicts the polynomial of the lane shape in an end-to-end manner based on the DETR structure [22]. BézierLaneNet uses the classic cubic Bézier curve to model the line shape, and the training process minimizes the distance between the ground truth curve and the predicted lane lines [5]. The advantage of BézierLaneNet is that the model converges easily and is less sensitive to hyperparameters.

## 3 Method

Despite extensive research across segmentation-based, anchor-based, and curve-based lane detection methods, robust lane detection in real-world scenarios remains a significant challenge. Key limitations include representational misalignment in segmentation approaches, strong dependence on manually designed, dataset-specific anchors in anchor-based methods, and instability caused by high-order polynomial coefficients in curve-based models, which can unpredictably distort lane geometry.
To address these issues, we propose a unified, end-to-end optimization strategy applicable to all three categories of lane detection models. Our approach is designed with the following key objectives:
- **Lightweight integration:** The optimization method should require minimal modifications to the original network architecture and avoid significantly increasing the number of parameters, ensuring compatibility with resource-constrained embedded systems.
- **Context-aware learning:** The method should possess a large receptive field to effectively capture non-local contextual features, enabling the detection of occluded or discontinuous lanes while maintaining clear separation between different lane instances.
- **Efficiency and accuracy:** The optimization technique must be simple to implement and capable of maintaining high inference speed while improving lane detection accuracy, thus supporting real-time deployment in low-power environments.

### 3.1 Covariance matrix and attention correlation

The overall model architecture is shown in Fig. 1. The input of CDO is an intermediate feature map $F \in R^{L \times H \times W}$. Each channel of $F$ can be viewed as a two-dimensional matrix $S \in H \times W$. In the training dataset, each input image has a binary lane mask. This mask is then scaled into a new mask $M_n$, which has dimensions of $H \times W$, using OpenCV's resize function with NEAREST interpolation. To calculate the covariance matrices of the horizontal and vertical lane features and masks, we measure the covariance between each row in matrix $S$ and each row in $M_n$ for the horizontal covariance matrix, and between each column in $S$ and each column in $M_n$ for the vertical covariance matrix. These two matrices are defined as:

$$COV_n^\theta = \begin{bmatrix} s_{11} & s_{12} & \cdots & s_{1W} \\ s_{21} & s_{22} & & s_{2W} \\ \vdots & \vdots & \ddots & \vdots \\ s_{H1} & s_{H2} & \cdots & s_{HW} \end{bmatrix} \begin{bmatrix} m_{11} & m_{12} & \cdots & m_{1W} \\ m_{21} & m_{22} & & m_{2W} \\ \vdots & \vdots & \ddots & \vdots \\ m_{H1} & m_{H2} & \cdots & m_{HW} \end{bmatrix}^T = \begin{bmatrix} h_{11} & h_{12} & \cdots & h_{1H} \\ h_{21} & h_{22} & & h_{2H} \\ \vdots & \vdots & \ddots & \vdots \\ h_{H1} & h_{H2} & \cdots & h_{HH} \end{bmatrix} \quad (1)$$

$$COV_n^\phi = \begin{bmatrix} s_{11} & s_{12} & \cdots & s_{1W} \\ s_{21} & s_{22} & & s_{2W} \\ \vdots & \vdots & \ddots & \vdots \\ s_{H1} & s_{H2} & \cdots & s_{HW} \end{bmatrix}^T \begin{bmatrix} m_{11} & m_{12} & \cdots & m_{1W} \\ m_{21} & m_{22} & & m_{2W} \\ \vdots & \vdots & \ddots & \vdots \\ m_{H1} & m_{H2} & \cdots & m_{HW} \end{bmatrix} = \begin{bmatrix} v_{11} & v_{12} & \cdots & v_{1W} \\ v_{21} & v_{22} & & v_{2W} \\ \vdots & \vdots & \ddots & \vdots \\ v_{W1} & v_{W2} & \cdots & v_{WW} \end{bmatrix} \quad (2)$$

Here $S$ and $M_n$ do matrix multiplication, where $n$ denotes the $n^{th}$ lane in the image, and $s_{i,j}$ is an item in matrix $S$, $m_{i,j}$ is an item in matrix $M_n$. $H$ and $W$ refer to the size of the feature map, which corresponds to the size of the horizontal and vertical covariance matrices, respectively. The $h_{ij}$ in $COV_n^\theta$ represents the relationship between the row vectors in matrix $S$ and the row vectors in $M_n$. The $v_{i,j}$ in vertical covariance matrix $COV_n^\phi$ represents the relationship between the column vectors in matrix $S$ and $M_n$. $\theta$, $\phi$ represent the horizontal and vertical directions, respectively. The horizontal covariance matrix measures the similarity of lane features and labels in the horizontal direction. It is more sensitive to changes occurring in this direction. Similarly, the vertical covariance matrix measures the similarity of lane features and labels in the vertical direction, and is more sensitive to changes in this direction. By analyzing both matrices, we can gain a more complete picture of changes in the spatial distribution of lanes and make more accurate predictions about the location of lane lines. The diagonal items in the covariance matrix can indicate the similarity in row or column and the off-diagonal items represent their correlation. The corresponding lane features in the feature map are enhanced by covariance calculation, while non-correlated information is suppressed.

In order to observe the saliency of diagonal items in covariance matrix, we define some simple function.

$$AVG(COV_n^\theta) = \frac{1}{H \times H} \sum_{i,j} h_{i,j} \quad (3)$$

$$AVG(COV_n^\phi) = \frac{1}{W \times W} \sum_{i,j} v_{i,j} \quad (4)$$

$$DIAG(COV_n^\theta) = \frac{1}{H} \sum_{i=j} h_{i,j} \quad (5)$$

$$DIAG(COV_n^\phi) = \frac{1}{W} \sum_{i=j} v_{i,j} \quad (6)$$

Here, AVG(·) is the average function to calculate mean value of the matrix. The DIAG(·) is the function to calculate the mean value of diagonal items of covariance matrix.

In the CDO process, all computational steps can be done with simple mathematical calculations. First, we calculate AVG(·) and DIAG(·) of the covariance matrix. Then, we use a relative intensity function (RIF) which is defined in Equations (7) and (8) to calculate the saliency of the diagonal items in the matrix. Finally, we compute the CDO output as the mean values of RIF across all channels. This output can be thought of as a quantitative measurement of the lane's saliency.

$$RIF_\theta(n) = \frac{|DIAG(COV_n^\theta) - AVG(COV_n^\theta)|}{\max(DIAG(COV_n^\theta), AVG(COV_n^\theta))} \tag{7}$$

$$RIF_\phi(n) = \frac{|DIAG(COV_n^\phi) - AVG(COV_n^\phi)|}{\max(DIAG(COV_n^\phi), AVG(COV_n^\phi))} \tag{8}$$

$$CDO(n) = \frac{1}{2C}\left[\alpha \sum_{i=1}^{C} RIF_\theta(n,i) + \beta \sum_{i=1}^{C} RIF_\phi(n,i)\right] \tag{9}$$

Here $RIF_\theta(n)$ and $RIF_\phi(n)$ calculate the relative intensity of the $n^{th}$ lane from the horizontal covariance and vertical covariance, respectively. In Equation (9), $CDO(n)$ computes the average RIF value for the $n^{th}$ lane. $\alpha$ and $\beta$ denote the weight coefficients in the horizontal and vertical directions, respectively. We can set $\alpha+\beta=1$, and the value of $\alpha$ and $\beta$ depends on the direction of the main lane in the detection. A typical value for α and β is α=0.5 and β=0.5. C represents the number of channels.

### 3.2 Add CDO to model

When designing a real-time lane detection model for a real application, it is important to keep in mind the importance of processing speed and minimal computing resources. As an attention mechanism only depend on intermediate feature output, CDO can be applied to various models with different structures. The intuition behind CDO is to propose an easier-to-use approach that focuses on mining long-range dependencies of features and obtaining global feature relationships in the horizontal and vertical directions through covariance calculations. Unlike non-local methods that require a lot of memory, CDO only uses matrix multiplication and requires very limited computing resources.

We have selected four SOTA lane detection models and two widely used real-time models to verify the effectiveness of CDO (Table 1). The four SOTA models are RESA, LaneATT, CLRNet, and BézierLaneNet. The two real-time models, ENet and ERFNet, are compared as baseline methods [23][24]. Table 1 shows the layer position of CDO in different networks. We can add CDO to different models in three steps. First, find the appropriate layer as the CDO to plug in. Then, select the feature map of $F$ of this layer as the CDO's input. The vertical and horizontal covariance calculations in Equations (1) and (2) can be simply done by matrix multiplication. The horizontal covariance matrix is $H \times H$, and the vertical covariance matrix is $W \times W$. Second, the RIF of the covariance matrix is calculated by Equations (7) and (8) to obtain the relative intensity of lane features in feature maps. Third, calculate the mean RIF values along the horizontal and vertical axes over all channels to define a criterion measuring the mean squared error (MSE) loss between the CDO output and lane existence exist(n), as shown in Equation (10).

$$Loss(CDO) = MSELoss(CDO(n), exist(n)) \tag{10}$$

Where the lane existence function exist(n) outputs the value of 0,1 for whether the $n^{th}$ lane exists.

### 3.3 Overall Loss

Table 1 Layer position of CDO in different networks

| Name | CDO Position | Category | Year |
|---|---|---|---|
| ENet | Layer 8 | segmentation-based | 2016 |
| ERFNet | Layer 16 | segmentation-based | 2018 |
| RESA (ResNet-34) | Layer 16 | segmentation-based | 2020 |
| LaneATT (ResNet-34) | Layer 16 | anchor-based | 2021 |
| CLRNet (ResNet-34) | Layer 16 | anchor-based | 2022 |
| BézierLaneNet (ResNet-34) | Layer 16 | curve-based | 2022 |

Table 2 Loss of CDO

| Name | Total Loss | Total Loss with CDO |
|---|---|---|
| ENet | | |
| ERFNet | $f_{seg} \cdot L_{seg} + f_{exist} \cdot L_{exist}$ | $f_{seg} \cdot L_{seg} + f_{exist} \cdot L_{exist} + f_{CDO} \cdot L_{CDO}$ |
| RESA (ResNet-34) | | |
| LaneATT (ResNet-34) | $f_{reg} \cdot L_{reg} + f_{cls} \cdot L_{cls}$ | $f_{reg} \cdot L_{reg} + f_{cls} \cdot L_{cls} + f_{CDO} \cdot L_{CDO}$ |
| CLRNet (ResNet-34) | $f_{cls} \cdot L_{cls} + f_{xytl} \cdot L_{xytl} + f_{seg} \cdot L_{seg} + f_{IoU} \cdot L_{IoU}$ | $f_{cls} \cdot L_{cls} + f_{xytl} \cdot L_{xytl} + f_{seg} \cdot L_{seg} + f_{IoU} \cdot L_{IoU} + f_{CDO} \cdot L_{CDO}$ |
| BézierLaneNet (ResNet-34) | $f_{cls} \cdot L_{cls} + f_{reg} \cdot L_{reg} + f_{seg} \cdot L_{seg}$ | $f_{cls} \cdot L_{cls} + f_{reg} \cdot L_{reg} + f_{seg} \cdot L_{seg} + f_{CDO} \cdot L_{CDO}$ |

Table 3 Loss coefficients of 6 models in three datasets

| Name | $f_{seg}$ | $f_{exist}$ | $f_{cls}$ | $f_{reg}$ | $f_{xytl}$ | $f_{IoU}$ | $f_{CDO}$ |
|---|---|---|---|---|---|---|---|
| ENet | 1 | 0.1 | - | - | - | - | 0.1 |
| ERFNet | 1 | 0.1 | - | - | - | - | 0.1 |
| RESA (ResNet-34) | 1 | 0.1 | - | - | - | - | 0.1 |
| LaneATT (ResNet-34) | - | - | 10 | 1 | - | - | 0.3 |
| CLRNet (ResNet-34) | 1 | - | 2/6/2 | - | 0.2/0.5/0.5 | 2/2/8 | 0.2 |
| BézierLaneNet (ResNet-34) | 0.75 | - | 0.1 | 1 | - | - | 0.05 |

The overall training loss of different models consists of different parts of the loss terms. We keep the auxiliary lane existence loss in all segmentation-based method [5] and segmentation loss in CLRNet [20]. As shown in Table 2, we enumerate how the losses from CDO are added to the total loss function of the six different models. The parameters $f_{seg}$, $f_{reg}$, $f_{exist}$, $f_{cls}$, $f_{IoU}$, $f_{xytl}$ and $f_{CDO}$ are the coefficients of segmentation losses ($L_{seg}$), regression loss ($L_{reg}$), existence loss ($L_{exist}$), classification loss ($L_{cls}$), line IoU loss ($L_{IoU}$), start point coordinate loss ($L_{xytl}$) and CDO loss ($L_{CDO}$), respectively. These parameters are used to balance the effect of different losses on the total loss.

After adding CDO to the total loss, the model can continue training from any epoch with the existing hyperparameters. By minimizing the loss of CDO, the features of the lanes will be enhanced through the information within a global context. In Table 3, it shows the loss coefficients for six models in three different datasets (CULane, TuSimple, and LLAMAS). To eliminate the dependence of CDO on datasets, the same $f_{CDO}$ value is used across all datasets.

### 3.4 Lane prediction

In practice, optimizing lane prediction using CDO is straightforward, and CDO only changes the training process. All image pre-processing, main program framework, and post-processing can remain unchanged. The hyperparameters optimized by the CDO retraining process can directly replace the original model files in real applications without worrying about code modifications and hardware support. Therefore, the lane inference process does not need to be changed.

## 4 Experiments

### 4.1 Dataset

In the experiment, three widely used lane detection datasets, CULane [12], TuSimple [25], and LLAMAS [26], were utilized to evaluate the performance changes of six models before and after adding CDO.

### 4.2 Evaluation metrics

For each distinct dataset, we employ the evaluation metrics listed in the literature to determine the effectiveness of CDO.

**CULane**. We employ 30-pixel-wide lines to connect the predicted and labeled lane points, respectively. We then calculate the IoU between the predicted lane line and the labeled lane line. True positives (TP) are the predicted lanes with an IoU above a certain threshold (0.5). The F1 measure is defined as: $F_1 = \frac{2 \cdot Precision \cdot Recall}{Precision + Recall}$. where $Precision = \frac{TP}{TP+FP}$ and $Recall = \frac{TP}{TP+FN}$. FP and FN are false positives (FP) and false negatives (FN), respectively.

**TuSimple**. For the TuSimple dataset, the evaluation metric is accuracy, and points are evaluated at a particular image height. Accuracy is defined as: $Accuracy = \frac{\Sigma_{clip} C_{clip}}{\Sigma_{clip} S_{clip}}$, where $C_{clip}$ is the number of correct points in the frame, $S_{clip}$ is the number of requested points. The predicted lane is referred to as a true positive (TP) if the difference between it and the ground truth point is less than a threshold. For evaluation, we also calculate the ratio of false positives (FP) and false negatives (FN) in Equation (11). FP denotes that the predicted lane has not been matched with any ground-truth lanes. FN denotes that no lane in the prediction matches the lane in the ground truth.

$$FP = \frac{F_{pred}}{N_{pred}}, FN = \frac{M_{pred}}{N_{gt}} \quad (11)$$

Where $F_{pred}$ is the number of incorrectly predicted lanes, $N_{pred}$ is the total number of predicted lanes. $M_{pred}$ is the number of missed ground-truth lanes in the predictions, $N_{gt}$ is the total number of ground-truth lanes.

**LLAMAS**. As a new large-scale lane detection dataset, LLAMAS contains over 100k images. BOSCH uses high-resolution maps to automatically generate annotations for publishing lane datasets, which makes this dataset one of the largest high-quality datasets that is freely available. We upload the lane predictions to the website of LLAMAS to get benchmark results for the test set.

### 4.3 Implementation details

In order to compare the results using the same software framework, we adopted the mmcv-based framework provided by Feng [26]. For preprocessing, the input resolutions are $288 \times 800$ for CULane and $360 \times 640$ for TuSimple and LLAMAS, respectively. To explore the effectiveness of CDO itself, we did not introduce any data augmentation. After applying CDO to six models (Table 4), we used the aforementioned three datasets to evaluate the optimized models. In order to make a fair comparison of the six models, both in their original form and with the addition of the CDO module, we trained them using the same optimizer, learning rate, and number of epochs. The coefficients of $\alpha$ and $\beta$ in Equation (9) are assigned a value of 0.5. Table 4 shows the configurations used in the experiments, which included different numbers of training epochs for the three datasets. For example, "12/50/10" indicates that CULane was trained for 12 epochs, TuSimple was trained for 50 epochs, and LLAMAS was trained for 10 epochs. All training and testing were conducted on a single RTX 1080 Ti. The computer specifications include a Xeon(R) Gold 6132 CPU @ 2.60GHz and 128GB of RAM.

### 4.4 Main results

Table 4 Training configurations and parameters

| Model | Opt. | LR schedule | Epochs |
| --- | --- | --- | --- |
| ENet | SGD | Poly policy with a warmup | 12/50/10 |
| ERFNet | SGD | Poly policy with warmup | 12/50/10 |
| RESA | SGD | Poly policy with warmup | 12/50/10 |
| LaneATT | Adam | Cosine Annealing | 15/100/15 |
| CLRNet | Adam | Cosine Annealing | 15/70/20 |
| BézierLaneNet | Adam | Cosine Annealing | 36/400/20 |

Table 5 Performance of different models with CDO on the CULane testing set

| Model<br>Proportion | Normal<br>27.70% | Crowded<br>23.40% | Night<br>20.30% | No line<br>11.70% | Shadow<br>2.70% | Arrow<br>2.60% | Curve<br>1.20% | Crossroad<br>9.00% | F1@50 |
|---|---|---|---|---|---|---|---|---|---|
| ENet | 89.27% | 68.15% | 62.98% | 42.42% | 68.65% | 83.07% | 63.23% | 2465 | 69.9% |
| ENet-CDO | 90.13% | 70.57% | 65.74% | 43.58% | 67.71% | 84.32% | 67.59% | 2400 | 71.66% ↑ |
| ERFNet | 91.48% | 71.27% | 68.07% | 46.76% | 74.5% | 86.12% | 66.97% | 2103 | 73.49% |
| ERFNet-CDO | 92.31% | 73.58% | 70.26% | 47.84% | 74.77% | 87.93% | 68.6% | 2472 | 74.93% ↑ |
| RESA (ResNet-34) | 91.32% | 71.81% | 67.55% | 46.53% | 72.64% | 86.97% | 67.22% | 1700 | 73.66% |
| RESA-CDO (ResNet-34) | 92.31% | 72.7% | 70.07% | 47.26% | 76.32% | 88.89% | 69.31% | 1679 | 75.08% ↑ |
| LaneATT (ResNet-34) | 91.34% | 73.72% | 70.42% | 48.4% | 73.66% | 86.8% | 66% | 964 | 75.8% |
| LaneATT-CDO (ResNet-34) | 91.6% | 73.59% | 71.2% | 48.2% | 71.47% | 86.83% | 68.43% | 992 | 75.89% ↑ |
| CLRNet (ResNet-34) | 93.35% | 78.1% | 74.94% | **52.92%** | **79.48%** | 90.22% | 70.46% | 1068 | 79.6% |
| CLRNet-CDO (ResNet-34) | **93.38%** | **78.12%** | **74.95%** | 52.28% | 79.43% | **90.51%** | **70.72%** | 1150 | **79.61%** ↑ |
| BézierLaneNet (ResNet-34) | 91.67% | 72.99% | 70.15% | 48.09% | 75.74% | 87.13% | 62.05% | 897 | 75.55% |
| BézierLaneNet-CDO(ResNet-34) | 91.69% | 73.07% | 70.27% | 48.31% | 75.4% | 86.98% | 62.51% | **856** | 75.65% ↑ |

The CDO can be added to the network without modifying the original structure. Commonly chosen layer options include the last layer of the encoder for segmentation-based models, or the last layer of the backbone for anchor-based and curve-based models. After the model is optimized by CDO, the updated parameter file can be directly used in the existing system. Six highly efficient real-time models were compared, including two lightweight models such as ENet and ERFNet, and four SOTA models such as RESA, LaneATT, CLRNet, and BézierLaneNet. After integrating the CDO module, the detection results significantly surpassed those based on the original structure. Considering ENet and ERFNet have only 1%-10% of the parameters of other SOTA methods, they are more suitable for mobile devices with limited memory resources.

**4.4.1 Performance of CDO on different datasets**

The CDO is implemented as a standalone module that can be integrated into various models' network structures.
**CULane**. Table 5 shows the performance of adding CDO to six different models on the testing set of CULane. For all categories of road scenes, after being optimized by the CDO module, all models achieved better performance compared to their original form. Among them, ENet-CDO achieved the most significant performance improvement of 1.8%. Additionally, the category of segmentation-based methods has better results than the other two categories.
**TuSimple**. Table 6 shows the performance of adding CDO to six different models on the testing set of TuSimple. After CDO optimization, the accuracy of the lane detection for most models has improved. Although the accuracy of LaneATT decreased, FP and FN have been significantly improved, which shows that the CDO is still effective in reducing false predictions. Among these six models, RESA-CDO has gained most improvement, and the accuracy has increased by 0.32%.
**LLAMAS**. The LLAMAS dataset restricts testing on four center lines and has 100,042 labeled images. We evaluated the CDO module on its validation and test set. The validation set was running on the local server, using the 0.5 and 0.75 IoU thresholds, respectively, and the test set was run on the official test server [26]. Table 7 shows the results of the model optimized through the CDO. In most cases, the F1 scores have increased. ENet-CDO has obtained the most improvement, with a 7.07% increase in F1@75, while LaneATT-CDO has decreased by 0.26% in F1@75.
As shown in Fig. 2, F1 improvements are accumulated on three datasets for six different models. It shows that CDO can significantly improve the model's detection results for almost all lane detection categories on all three datasets. For segmentation-based methods, it reaches about 1.7-2.4% increasement. For the anchor-based methods, it reaches about 0.1-0.5% F1 increasement. For curve-based methods, it reaches about 0.5% performance increasement.

Table 6 Performance of different models with CDO on TuSimple testing set

| Model | Accuracy | FP | FN |
|---|---|---|---|
| ENet | 95.62% | 6.54% | 5.02% |
| ENet-CDO | 95.72% ↑ | 5.85% | 4.68% |
| ERFNet | 96.03% | 5.91% | 3.65% |
| ERFNet-CDO | 96.09% ↑ | 6.57% | 3.59% |
| RESA (ResNet-34) | 95.15% | 6.92% | 5.92% |
| RESA-CDO (ResNet-34) | 95.47% ↑ | 5.66% | 5.36% |
| LaneATT (ResNet-34) | 95.46% | 12.77% | 3.26% |
| LaneATT-CDO (ResNet-34) | 95.25% ↓ | 7.93% | 3.09% |
| CLRNet (ResNet-34) | 96.8% | 2.18% | 2.08% |
| CLRNet-CDO (ResNet-34) | 96.83% ↑ | 2.21% | 2.08% |
| BézierLaneNet (ResNet-34) | 95.65% | 5.15% | 3.87% |
| BézierLaneNet-CDO (ResNet-34) | 95.75% ↑ | 3.91% | 3.46% |

This result shows that CDO is more effective for segmentation-based methods than anchor-based or curve-based methods. We suppose that by minimizing the CDO loss described in Equation (10), the segmentation-based methods are more likely to maintain global information by enforcing the spatial similarity between the feature map and the label mask during the feature extraction process. After the network parameter is optimized by CDO, the spatial relationship perception ability of the model will be enhanced, thereby increasing the overall lane identification capability. Although CDO is less effective in anchor-based and curve-based networks, it still helps these networks to improve performance by extracting spatial level correlation information at a high semantic level.

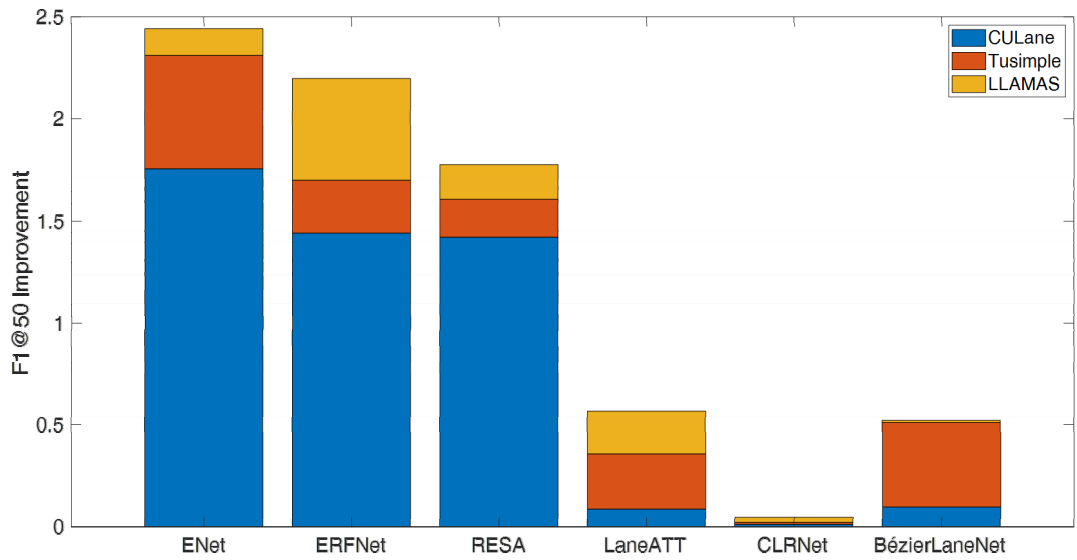

**Fig. 2** The variation in F1 scores across three datasets for different models after CDO optimization. Different colors represent the improvement in F1 scores for different datasets, and the total height of the bar represents the overall improvement effect of each model.

Table 7 Performance of different models with CDO on LLAMAS test server

| Model | Valid | Test | |
|---|---|---|---|
| | F1@50 | F1@75 | F1@50 |
| ENet | 96.13% | 73.52% | 94.43% |
| ENet-CDO | 96.26% ↑ | 80.59% ↑ | 95.5% ↑ |
| ERFNet | 95.83% | 76.65% | 94.8% |
| ERFNet-CDO | 96.33% ↑ | 80.73% ↑ | 95.21% ↑ |
| RESA (ResNet-34) | 96.28% | 76.93% | 95.32% |
| RESA-CDO (ResNet-34) | 96.45% ↑ | 80.15% ↑ | 95.53% ↑ |
| LaneATT (ResNet-34) | 90.45% | 60.09% | 93.74% |
| LaneATT-CDO (ResNet-34) | 90.68% ↑ | 59.83% ↓ | 93.79% ↑ |
| CLRNet (ResNet-34) | 96.48% | 84.98% | 96.08% |
| CLRNet-CDO (ResNet-34) | 96.51% ↑ | 84.99% ↑ | 96.15% ↑ |
| BézierLaneNet (ResNet-34) | 93.63% | 64.23% | 94.91% |
| BézierLaneNet-CDO(ResNet-34) | 93.64% ↑ | 64.46% ↑ | 94.96% ↑ |

Table 8 Inference speed (FPS) and parameters comparison of different models with CDO

| Model | Parameter (M) | MAdd | GFLOPs | Memory (MB) | Inference Speed (FPS) |
|---|---|---|---|---|---|
| ENet-CDO | 0.35 | 3.68 | 1.81 | 378 | 69 |
| ERFNet-CDO | 2.06 | 23.66 | 11.47 | 343 | 110 |
| RESA-CDO (ResNet-34) | 11.39 | 96.81 | 47.69 | 651 | 62 |
| LaneATT-CDO (ResNet-34) | 22.12 | 34.39 | 18.02 | 174 | 165 |
| CLRNet-CDO (ResNet-34) | 21.87 | 38.04 | 20.79 | 199 | 88 |
| BézierLaneNet-CDO (ResNet-34) | 8.90 | 28.69 | 14.36 | 176 | 185 |

### 4.4.2 Comparison with state-of-the-art methods

We choose ResNet-34 as the backbone for RESA, LaneATT, CLRNet and BézierLaneNet to ensure that all SOTA methods are compared under the same conditions. For segmentation-based methods, such as RESA, the CDO is integrated at the last layer of the encoder of the model. For anchor-based and curve-based methods, the CDO insertion layer can be the last layer of the backbone output (the details are shown in Table 1).Table 5, 6, and 7 show the results of adding the CDO to the baseline method and SOTA method. The areas marked by yellow rectangles in Fig. 3 show that CDO produces positive enhancements to lane detection in all categories of lane detection models. Unlike many segmentation-based methods that require redesigning the encoder, CDO does not introduce additional structure to the model, it just adds the CDO loss function to the original network. Meanwhile, CDO does not rely on any pre-designed structure, such as anchor. During the experiments, we evaluated the performance of five key indices on a single RTX 1080 Ti, which included parameters, multiplications and additions (MAdd), GFLOPs, memory, and FPS. Since CDO only increases computation time during training, these metrics for the models with and without CDO do not change during the inference process. According to the results, all networks achieve real-time performance (>50fps) as shown in Table 8.

Table 9 Comparison of the F1 of SAD and CDO on the CULane dataset

| Model | Original | SAD | CDO |
|---|---|---|---|
| ENet | 69.9% | 70.8% | 71.66% |
| ERFNet | 73.49% | 73.61% | 74.67% |
| MobileNetV2 | 67.41% | 67.40% | 68.77% |
| DeepLabV1 (ResNet-34) | 69.92% | 70.41% | 70.9% |
| SCNN (ResNet-34) | 72.70% | 72.61% | 74.01% |
| RESA (ResNet-34) | 73.66% | 73.50% | 75.08% |

## 5 Ablation study

### 5.1 SAD and CDO

Self-attention distillation (SAD) allows a model to learn from itself and gains substantial improvement without any additional supervision or labels [27]. Other advantages of SAD include its easy incorporation into any feed-forward convolutional neural network (CNN) without increasing inference time. Therefore, we selected SAD in this paper to compare with CDO and observe the optimization effect of these two different attention modules. Since SAD is only applicable to segmentation-based models, we chose six segmentation-based methods to add CDO and SAD for comparison to explore their effectiveness in lane detection. In the experiment, we assumed a half-trained model and then added SAD or CDO into the training process.

As can be seen in Table 9, CDO brings more performance gains than SAD in all six models. We attribute this to the following causes. First, CDO makes it possible for the network to be more aware of global relationships, which helps CDO outperform SAD in terms of performance. Second, while using attention maps to distill contextual information for lane detection, the choice of SAD depends on the network's structure and is typically only appropriate for feed-forward CNN architectures. Third, SAD may have possible conflicting influence with other attention such as recurrent or spatial CNN, but CDO can be compatible with different attention modules.

### 5.2 Ablation study on covariance distribution

After the CDO training process, the improvement of the covariance distribution will change the spatial accuracy of the lane features in the feature map. When the relative strength of the diagonal elements in the covariance matrix increases, the lane prediction accuracy also increases, and the result is shown in Fig. 5(b). In the CDO re-training process, the RIF described in Equation (7) has increased significantly.

When the RIF increases with the training epoch, the CDO's attention mechanism makes the features of the lane clearer in context. With the increase in correlation between the feature of the lane and its actual position, false positive and false negative predictions will also be reduced.

The misclassification of the lane prediction point (in Fig. 5(a) the red dashed line) is also reduced. After adding the CDO to the total loss, the covariance distribution seems to inhibit non-lane features and reduce the interference of irrelevant information in lane prediction, and the model does not require any post-processing steps.

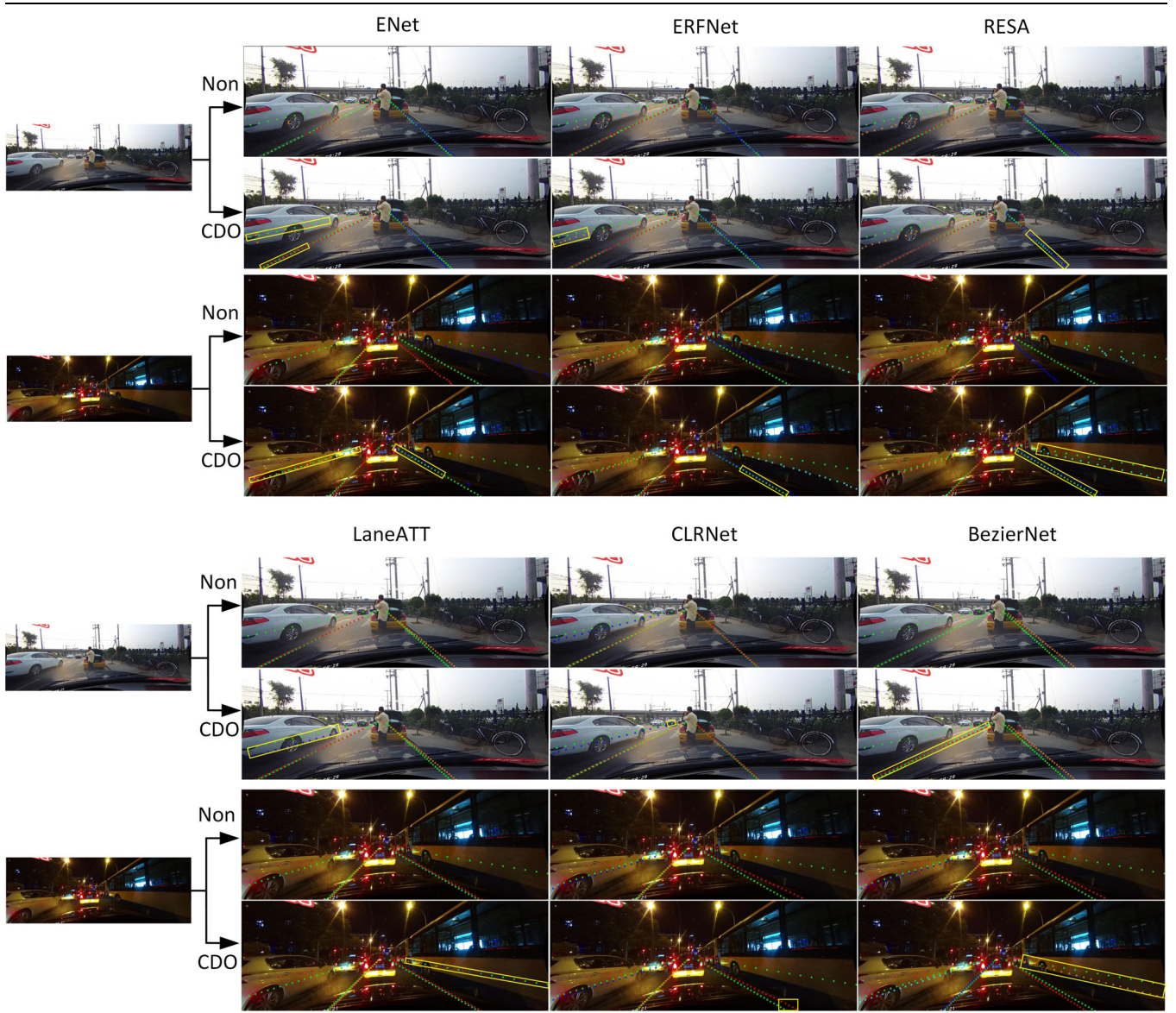

**Fig. 3** Visualization results of different models with and without CDO on CULane testing set. The green points are the ground truth, other color indicate different lane instance. Yellow rectangular boxes are used in the figure to highlight areas where lane detection accuracy has improved.

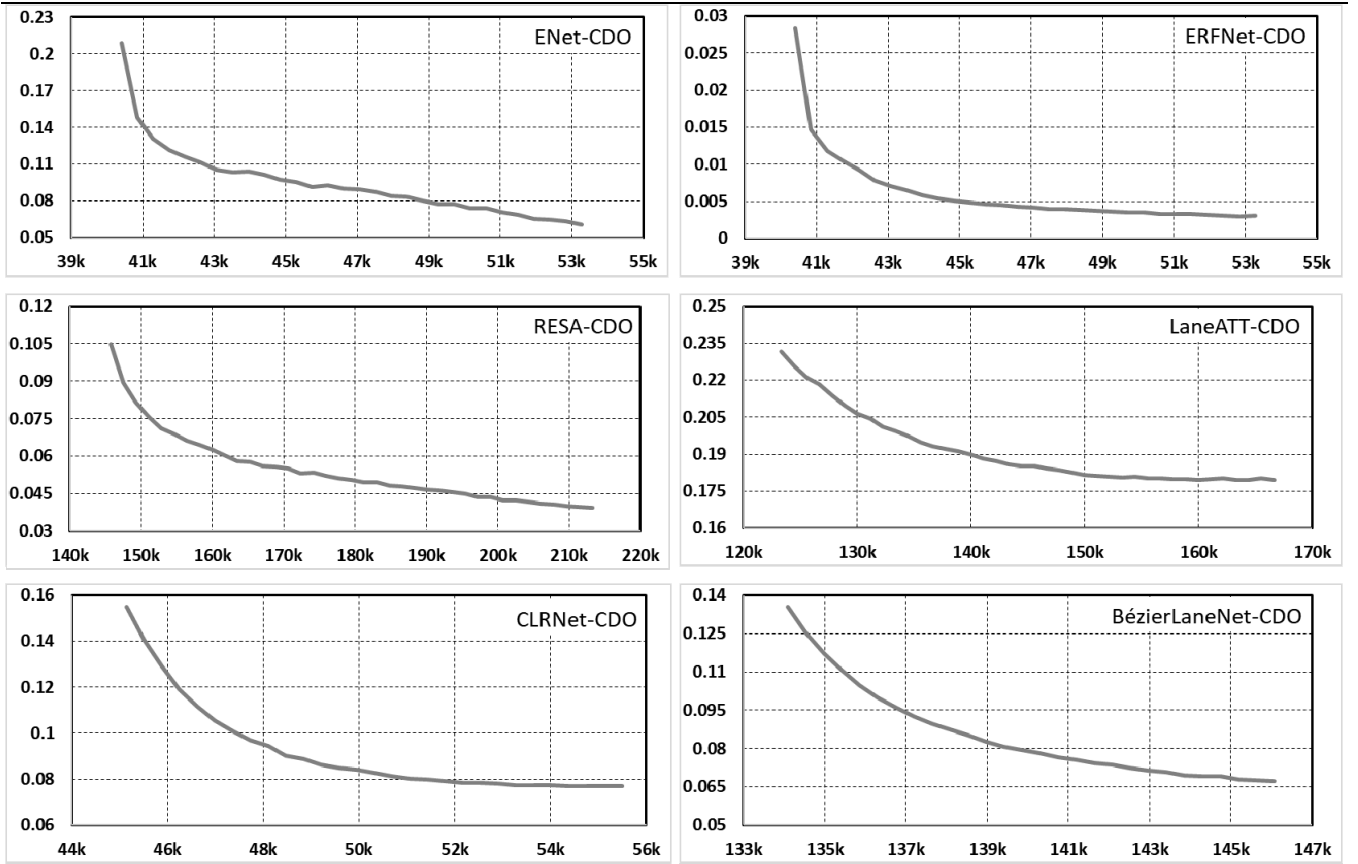

**Fig. 4** CDO loss convergence results in different models during training.

## 5.3 Ablation study on CDO loss

In the CDO training process, the CDO loss is a part of the total loss. In practical applications, the weight of CDO loss depends on the category of lane detection. Under the same category, the CDO loss coefficient in Table 2 can remain unchanged. When the CDO loss decreases, the trained model will focus more attention on the overall existence of the lane, the thin line shape of the lane, and the spatial accuracy of the lane. Before the optimization of CDO, the original network is more sensitive to the local features of the lane. After CDO, the network will focus more on the integrity of the overall line structure and the correctness of the lanes' spatial positions. Fig. 4 shows that after adding the CDO module to the network, the CDO loss decreases and converges as the training epoch increases. In the experiments, we added the CDO branch to the model after 75% of the total training process. After the CDO is inserted into the model and continues to train, the RIF in Equations (7) and (8) will increase with the training epochs (Fig. 5(c)), and the global lane feature extraction will also improve. In Fig. 5(c), the missing lane on the left side at epoch=10 will be detected in epoch=11 and 12. Similarly, the rightmost lane marking, which was erroneously detected due to a false positive, was correctly excluded. In addition, the spatial accuracy of the center lane is significantly improved.

In addition, we also studied the impact of different $f_{CDO}$ on model performance. In order to overcome the impact of other hyperparameter changes on the experimental results, we kept the hyperparameters in Table 3 unchanged during the experiment and observed changes in model performance by changing the value of $f_{CDO}$.

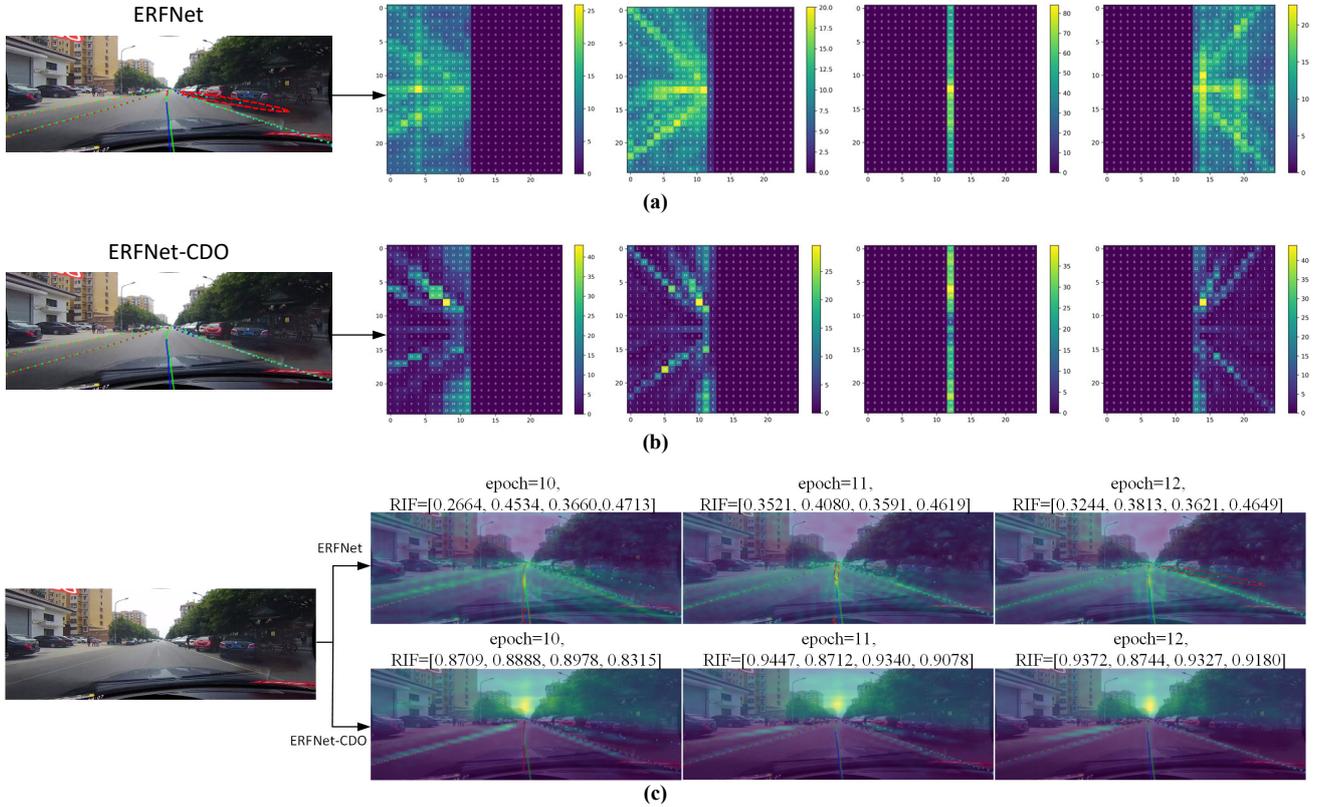

**Fig. 5** Comparison of covariance distribution between ERFNet and ERFNet-CDO.
(a) ERFNet's lane prediction is marked with colorful points. The ground truth lane is represented by green point. The red dotted line circle is an instance of a false positive lane. On the left is the covariance distribution map of four different lane lines.
(b) ERFNet-CDO's lane prediction is more accurate than ERFNet. The diagonal elements in the covariance matrix are more obvious than the covariance matrix of ERFNet.
(c) The visualization of the Grad-CAM of ERFNet-CDO. After the introduction of CDO into the re-training process, the RIF described in Equation (7) has significantly increased.

# 6  Conclusion

In this paper, we introduce a Covariance Distribution Optimization (CDO) approach to enhance the performance of real-time lane detection models, particularly in low-power embedded environments. The proposed CDO module leverages the covariance relationship between the feature map and ground-truth labels to construct a novel loss function. Specifically, it utilizes the diagonal elements of horizontally and vertically computed covariance matrices to enable each pixel to aggregate global contextual information, thereby improving spatial feature alignment.
We evaluate CDO across six representative models spanning the three main categories of lane detection methods—segmentation-based, anchor-based, and curve-based. In all cases, integrating CDO consistently leads to improved lane detection accuracy compared to the original model architectures. Notably, CDO does not require model retraining from scratch or the introduction of additional network parameters. Instead, it can be seamlessly incorporated into existing systems by adding the CDO loss term to the overall loss function and updating the model parameters accordingly.

Comprehensive experiments on three widely-used benchmark datasets—CULane, TuSimple, and LLAMAS—demonstrate the effectiveness of our approach, with consistent performance gains observed across all evaluated state-of-the-art models. In future work, we aim to extend the applicability of CDO to 3D lane detection frameworks and bird's-eye-view (BEV) representations, which would further broaden its relevance to advanced autonomous perception systems.

## Acknowledgment

This research was supported by Zhejiang Provincial Natural Science Foundation of China under Grant No. LGG21F030008.

## References

1  Sato, T., Chen, Q.A.: 'Towards Driving-Oriented Metric for Lane Detection Models', in 'Proceedings of the IEEE Computer Society Conference on Computer Vision and Pattern Recognition' (2022), pp. 17132–17141

2  Yurtsever, E., Lambert, J., Carballo, A., Takeda, K.: 'A Survey of Autonomous Driving: Common Practices and Emerging Technologies'*IEEE Access*, 2020, **8**, pp. 58443–58469.

3  Tang, J., Li, S., Liu, P.: 'A review of lane detection methods based on deep learning'*Pattern Recognit.*, 2021, **111**, p. 107623.

4  Bar Hillel, A., Lerner, R., Levi, D., Raz, G.: 'Recent progress in road and lane detection: A survey'*Mach. Vis. Appl.*, 2014, **25**, (3), pp. 727–745.

5  Feng, Z., Guo, S., Tan, X., Xu, K., Wang, M., Ma, L.: 'Rethinking Efficient Lane Detection via Curve Modeling'*Proc. IEEE Comput. Soc. Conf. Comput. Vis. Pattern Recognit.*, 2022, **2022-June**, pp. 17041–17049.

6  Liu, Y., Xu, P., Zhu, L., Yan, M., Xue, L.: 'Reinforced attention method for real-time traffic line detection'*J. Real-Time Image Process.*, 2022, **19**, (5), pp. 957–968.

7  Bardes, A., Ponce, J., LeCun, Y.: 'VICReg: Variance-Invariance-Covariance Regularization for Self-Supervised Learning'2021, (2020), pp. 1–23.

8  Zakaria, N.J., Shapiai, M.I., Ghani, R.A., Yassin, M.N.M., Ibrahim, M.Z., Wahid, N.: 'Lane Detection in Autonomous Vehicles: A Systematic Review'*IEEE Access*, 2023, **11**, (January), pp. 3729–3765.

9  Lee, D.H., Liu, J.L.: 'End-to-end deep learning of lane detection and path prediction for real-time autonomous driving'*Signal, Image Video Process.*, 2022, pp. 1–5.

10  Xiao, L., Li, X., Yang, S., Yang, W.: 'ADNet: Lane Shape Prediction via Anchor Decomposition'*Proc. IEEE/CVF Int. Conf. Comput. Vis.*, 2023, pp. 6404–6413.

11  Wang, Z., Ren, W., Qiu, Q.: 'LaneNet: Real-Time Lane Detection Networks for Autonomous Driving'2018.

12  Pan, X., Shi, J., Luo, P., Wang, X., Tang, X.: 'Spatial as deep: Spatial CNN for traffic scene understanding'*32nd AAAI Conf. Artif. Intell. AAAI 2018*, 2018, pp. 7276–7283.

13  Neven, D., De Brabandere, B., Georgoulis, S., Proesmans, M., Van Gool, L.: 'Towards End-to-End Lane Detection: An Instance Segmentation Approach'*IEEE Intell. Veh. Symp. Proc.*, 2018, **2018-June**, (June), pp. 286–291.

14  Zheng, T., Fang, H., Zhang, Y., *et al.*: 'RESA: Recurrent Feature-Shift Aggregator for Lane Detection'*35th AAAI Conf. Artif. Intell. AAAI 2021*, 2021, **4B**, pp. 3547–3554.

15  Sun, Y., Li, J., Xu, X., Shi, Y.: 'Adaptive Multi-Lane Detection Based on Robust Instance Segmentation for Intelligent Vehicles'*IEEE Trans. Intell. Veh.*, 2023, **8**, (1), pp. 888–899.

16  Li, X., Li, J., Hu, X., Yang, J.: 'Line-CNN: End-to-End Traffic Line Detection with Line Proposal Unit'*IEEE Trans. Intell. Transp. Syst.*, 2020, **21**, (1), pp. 248–258.

17  Tabelini, L., Berriel, R., Paixão, T.M., Badue, C., de Souza, A.F., Oliveira-Santos, T.: 'Keep your Eyes on the Lane: Real-time Attention-guided Lane Detection', in 'Proceedings of the IEEE Computer Society Conference on Computer Vision and Pattern Recognition' (2021), pp. 294–302

18  Qin, Z., Wang, H., Li, X.: 'Ultra Fast Structure-aware Deep Lane Detection'2020, pp. 1–16.

19  Liu, L., Chen, X., Zhu, S., Tan, P.: 'CondLaneNet: a Top-to-down Lane Detection Framework Based on Conditional Convolution'*Proc. IEEE Int. Conf. Comput. Vis.*, 2021, pp. 3753–3762.

20  Zheng, T., Huang, Y., Liu, Y., *et al.*: 'CLRNet: Cross Layer Refinement Network for Lane Detection'*Proc. IEEE Comput. Soc. Conf. Comput. Vis. Pattern Recognit.*, 2022, **2022-June**, (c), pp. 888–897.

21  Tabelini, L., Berriel, R., Paixão, T.M., Badue, C., de Souza, A.F., Oliveira-Santos, T.: 'PolyLaneNet: Lane estimation via deep polynomial regression'*Proc. - Int. Conf. Pattern Recognit.*, 2020, pp. 6150–6156.

22 Liu, R., Yuan, Z., Liu, T., Xiong, Z.: 'End-to-end lane shape prediction with transformers'*Proc. - 2021 IEEE Winter Conf. Appl. Comput. Vision, WACV 2021*, 2021, pp. 3693–3701.
23 Paszke, A., Chaurasia, A., Kim, S., Culurciello, E.: 'ENet: A Deep Neural Network Architecture for Real-Time Semantic Segmentation'2016, pp. 1–10.
24 Romera, E., Alvarez, J.M., Bergasa, L.M., Arroyo, R.: 'ERFNet: Efficient Residual Factorized ConvNet for Real-Time Semantic Segmentation'*IEEE Trans. Intell. Transp. Syst.*, 2018, **19**, (1), pp. 263–272.
25 Tusimple: 'TuSimple: Lane Detection Challenge (Dataset)' (2017)
26 Behrendt, K., Soussan, R.: 'Unsupervised labeled lane markers using maps'*Proc. - 2019 Int. Conf. Comput. Vis. Work. ICCVW 2019*, 2019, pp. 832–839.
27 Hou, Y., Ma, Z., Liu, C., Loy, C.C.: 'Learning lightweight lane detection CNNS by self attention distillation'*Proc. IEEE Int. Conf. Comput. Vis.*, 2019, **2019-Octob**, pp. 1013–1021.